\pdfoutput=1

\documentclass[11pt]{article}

\usepackage[final]{acl}

\usepackage{times}
\usepackage{latexsym}
\usepackage{amsmath}
\usepackage{placeins}  
\usepackage{tabularx}
\usepackage{algorithm}
\usepackage{algpseudocode}
\usepackage{booktabs}  

\usepackage[T1]{fontenc}

\usepackage[utf8]{inputenc}

\usepackage{microtype}

\usepackage{inconsolata}

\usepackage{graphicx}
\usepackage{float}

\title{MAGE: Multi-Head Attention Guided Embeddings for Low Resource Sentiment Classification}

\author{
 \textbf{Varun Vashisht\textsuperscript{1}},
 \textbf{Samar Singh\textsuperscript{1}},
 \textbf{Mihir Konduskar\textsuperscript{1}},
 \textbf{Jaskaran Singh Walia\textsuperscript{1}},
\\
 \textbf{Vukosi Marivate\textsuperscript{2}}
\\
 \textsuperscript{1}School of Computer Science and Engineering, Vellore Institute of Technology \\
 \textsuperscript{2}Department of Computer Science, University of Pretoria,
\\
 \small{
   \textbf{Correspondence:} \href{mailto:varunvashisht.work@gmail.com}{varunvashisht.work@gmail.com}
 }
}

\begin{document}
\maketitle
\begin{abstract}
Due to the lack of quality data for low-resource Bantu languages, significant challenges are presented in text classification and other practical implementations. In this paper, we introduce an advanced model combining Language-Independent Data Augmentation (LiDA) with Multi-Head Attention based weighted embeddings to selectively enhance critical data points and improve text classification performance. This integration allows us to create robust data augmentation strategies that are effective across various linguistic contexts, ensuring that our model can handle the unique syntactic and semantic features of Bantu languages. This approach not only addresses the data scarcity issue but also sets a foundation for future research in low-resource language processing and classification tasks.
\end{abstract}

\section{Introduction}
 Text classification is one of the most widely explored tasks in Natural Language Processing (NLP) due to its diverse applications, including spam detection, sentiment analysis, and topic modeling. Despite the impressive advancement achieved through deep learning, these methods rely heavily on large amounts of labeled data, posing a challenge for low-resource languages \citep{nie-etal-2023-cross, ogueji-etal-2021-small}. African languages in general, exemplify this challenge, as the scarcity of annotated datasets limits the development of effective text classification models \citep{amol2024statenlpkenyasurvey, Adelani2022}. Data augmentation has emerged as a promising solution for addressing data scarcity by generating synthetic data from original datasets \citep{kobayashi2018contextualaugmentationdataaugmentation}. Traditional augmentation techniques, including synonym replacement, sentence back-translation, and generative models, rely heavily on language-specific resources such as pre-trained word embeddings, language models, or linguistic databases, such as WordNet \citep{miller-1994-wordnet, wei2019edaeasydataaugmentation, percin-etal-2022-combining, Jahan2022}. This language dependence makes these approaches less effective for underrepresented languages, such as Bantu languages, which lack these linguistic resources \citep{10.1162/coli_a_00425}. To overcome these limitations, Language-Independent Data Augmentation (LiDA) was introduced \citep{LiDA}, which operates at the sentence embedding level rather than word or sentence levels. LiDA transforms sentence embeddings to generate synthetic data, bypassing the need for language-specific resources. Building upon this foundation, we propose MAGE (Multi-Head Attention Guided Embeddings), a framework designed to enhance text classification performance for low-resource languages. MAGE extends the LiDA framework by introducing significant innovations to the embedding and augmentation process. Specifically, it replaces the traditional Denoising Autoencoder with a Variational Autoencoder (VAE) to enable more expressive and diverse synthetic embeddings. Additionally, MAGE incorporates a novel Multi-Head attention mechanism that selectively emphasizes salient features in the embeddings. This focus on Multi-Head  attention improves the model’s capacity to capture critical syntactic and semantic nuances, making it particularly effective for low-resource languages. Using the AfriSenti SemEval dataset \citep{muhammadSemEval2023}, a collection of tweets annotated with positive, negative, and neutral sentiments for Kinyarwanda, Swahili, and Tsonga, we evaluate the performance of MAGE in sentiment classification. Our results demonstrate that MAGE outperforms baseline approaches in low-resource settings. Moreover, comparative analyses highlight the advantages of MAGE over self-attention-based models, further establishing its value as a robust framework for addressing the challenges posed by data scarcity in low-resource languages. This work not only addresses the pressing issue of data scarcity in Bantu languages but also provides a scalable and adaptable framework for extending text classification capabilities to other low-resource language families. Through the introduction of MAGE, we set the stage for future research in low-resource language processing and establish a pathway to improve the inclusivity and generalizability of NLP technologies.

\section{Related Works}

\subsection{Data Augmentation Techniques}
In recent years, data augmentation techniques have gained significant attention, especially for low-resource languages, due to the scarcity of properly annotated datasets and general lack of resources. Analyzing one such technique by Sennrich et al. (\citeyear{sennrich2016improvingneuralmachinetranslation}) leverage monolingual target language data for textual-based data augmentation using back-translation to enhance model performance. However, this method requires extra computational resources, as it needs an additional pre-trained NMT model.

Another approach proposed by  Lample et al. (\citeyear{lample2018unsupervisedmachinetranslationusing}) relies solely on monolingual corpora, which eliminates the need for parallel data. This is achieved by mapping sentences from two languages into a shared latent space using a shared encoder-decoder architecture. Although this method avoids the need for parallel data, it still depends on monolingual corpora, which can be scarce for certain languages. For a simpler and highly effective solution, Wei and Zou (\citeyear{wei2019edaeasydataaugmentation}) introduced the "EDA: Easy Data Augmentation Techniques for Boosting Performance on Text Classification Tasks." Their technique consists of four operations: synonym replacement, random insertion, random swap, and random deletion. This straightforward method significantly boosted model performance, even when training with only half of the available dataset. EDA operates without dependencies beyond a basic synonym dictionary like Miller (\citeyear{miller-1994-wordnet}).

In \citeauthor{kobayashi2018contextualaugmentationdataaugmentation} (\citeyear{kobayashi2018contextualaugmentationdataaugmentation}), the authors proposed a novel approach called contextual augmentation. Unlike previous techniques that relied on predefined synonym dictionaries or rules, this model uses a bidirectional language model to predict appropriate substitute words based on the surrounding context. This allows for more diverse word suggestions and has shown promising potential, consistently outperforming previous techniques like EDA, especially in low-resource scenarios. In
\citep{chen2020mixtextlinguisticallyinformedinterpolationhidden} explored a technique called TMix, which interpolates hidden space representations of text samples to create continuous augmented data, helping to mitigate overfitting in resource-limited settings. Their study also incorporated entropy minimization and consistency regularization to effectively utilize both labeled and unlabeled data, showing great potential in scenarios with extremely limited labeled data.
Another innovative approach, proposed by \citeauthor{yu2017seqgansequencegenerativeadversarial} (\citeyear{yu2017seqgansequencegenerativeadversarial}), is SeqGAN, which combines reinforcement learning with Generative Adversarial Networks (GANs) to address challenges in generating discrete sequences. SeqGAN uses Monte Carlo search and models the generator as a stochastic policy to overcome the difficulty of propagating gradients through discrete outputs. The discriminator provides rewards for complete sequences, guiding the generator via policy gradient updates. This method demonstrated significant improvements over baseline methods in tasks like music generation and extends GANs for discrete sequence generation in NLP tasks. \citeauthor{jia2017adversarialexamplesevaluatingreading} (\citeyear{jia2017adversarialexamplesevaluatingreading}) analyzed an adversarial evaluation scheme for reading comprehension systems, targeting the SQuAD dataset \citeauthor{rajpurkar-etal-2016-squad} (\citeyear{rajpurkar-etal-2016-squad}). This method involves adding adversarially crafted sentences to input paragraphs, designed to confuse machine learning models while remaining clear to humans. The study showed that even state-of-the-art models experienced performance drops under adversarial evaluation, highlighting the importance of genuine language understanding.
Raffel et al. (\citeyear{raffel2023exploringlimitstransferlearning}) delved into transfer learning, enabling models trained on high-resource languages to be adapted for low-resource languages. They examined methods like sentence-level alignment and multilingual embeddings, which integrate linguistic resources across languages. These techniques can significantly improve model performance in low-resource scenarios by leveraging existing high-resource language data.
Lastly, Li et al. (\citeyear{li2023syntheticdatagenerationlarge}) explored generating synthetic data for text classification using large language models. They used zero-shot and few-shot settings to generate synthetic training data with tailored prompts and manual annotation. The study found that model performance with synthetic data is affected by the subjectivity of the classification task. Incorporating real-world examples and human feedback can enhance the quality of synthetic data and improve model performance, especially for subjective tasks.

\subsection{Data Augmentation for Low Resource Corpora Text Classification}
One of the problems performing various downstream tasks in NLP is working with languages that don't have much labeled data. At such times, data augmentation techniques help tackle the problem. One such approach is TAU-DR by Rahamim
et al. (\citeyear{rahamim-etal-2023-text}) which uses soft prompts and keeps language model frozen to reconstruct hidden representations and then turning those back into synthetic sentences and proves to be really effective at improving multi-class classification. Additionally \cite{thangaraj2024crosslingualtransfermultilingualmodels} also conducted a study on the cross-lingual transfer capabilities in low-resource african languages and benchmarked the forgetting metrics, although this study did not employ augmentation metrics to reduce the model's forgetting.
Some other techniques like AEDA in Karimi et al. (\citeyear{karimi-etal-2021-aeda-easier}) which inserts punctuation marks into sentences to create variations unlike methods like EDA where data loss is possible as it involves operations like deletion and substitution, this method preserves the original information and semantic consistency. After testing on multiple datasets AEDA consistently outperformed EDA, particularly in low-resource settings, The simplicity of AEDA and its performance across different models highlights its effectiveness. IndiText Boost by Litake et al. (\citeyear{Litake2024IndiTextBT}), is a framework tailored for the underrepresented Indian languages, which utilizes techniques like EDA and back-translation to outperform the more complex LLM-based methods in basic classification tasks.
 
 In Sahu et al. \citeyear{sahu-etal-2022-data}, the authors explore prompt-based data augmentation in intent classification tasks using Large Language Models (LLMs) like gpt2 \citep{radford2019language} which boils down to just using a pre-trained LLM to augment data by creating synthetic data but due to LLMs tendency to create unreliable data this can lead to reducing data's quality. In Zhao et al. (\citeyear{zhao-etal-2022-epida}), the authors proposed EPiDA, (Easy Plug-in Data Augmentation) framework, which works for text classification. It employs two methods, conditional entropy minimization and relative entropy maximization which balances the diversity and quality of augmented data. Conditional entropy minimization ensures the semantic consistency while relative entropy maximization promotes more diverse samples. Through extensive testing it was seen that it consistently outperforms existing data augmentation techniques in various NLP tasks highlighting its application for low-resource applications. 
\\
\subsection{LiDA - Language Indpendent Data Augmentation}

The LiDA: Language-Independent Data Augmentation for Text Classification by
 Sujana and Kao (\citeyear{LiDA}) introduced us to a novel augmentation technique catering to low-resource language settings. The authors, rather than creating synthetic data for increasing the dataset, worked on augmenting the sentence level embeddings for the classification task, resulting in a 2\%-3\% improvement on average in LSTM classification results. The language independence mentioned arises from the multilingual dataset that is used for the training of the SBERT model.

\section{Methodology}
To further refine the LiDA architecture, we propose a Multi-Head  attention-based mechanism to quantitatively highlight and weight the individual embeddings to emphasize the important contributions of the LiDA architecture for the text-classification goal. 

\subsection{Dataset}
The dataset referred to is the AfriSenti SemEval Shared Task - 12 dataset by Muhammad et al. (\citeyear{muhammadSemEval2023}) based on tweet sentiment analysis. As the study focuses on the Bantu language family, the datasets of the following 3 Bantu languages were chosen - Kinyarwanda, Xitsonga and Swahili having 7940 tweet-label pairs in the combined training set and 1482 tweet-label pairs in the combined test set.
We observe a skewness in the data towards Kinyarwanda due to Kinyarwanda having the highest data points at 5155 tweets, with Swahili being the second highest at 3009 tweets and Xitsonga having the least data at 1258 tweets.

\begin{table}[h]
  \centering
  \begin{tabular}{lc}
    \hline
    \textbf{Field} & \textbf{Description} \\
    \hline
    ID     & Alpha-Numeric Serial Numbers \\
    Tweet    & Tweet Content \\
    Label     & Tweet Label \\
    \hline
  \end{tabular}
\caption{Field descriptions of the dataset.}
  \label{tab:dataset}
\end{table}

The dataset consists of 3 main fields, namely ID, Tweet, and Label. On preprocessing the data, the ID field was dropped, and the tweets accordingly preprocessed: lower casing and removing punctuations, hyperlinks, and emojis. The labels in the label field as mentioned in table~\ref{tab:dataset} were given as Negative, Neutral and Positive to indicate the tweet sentiment which was label-encoded as 0,1,2 respectively for further processing.

\subsection{Architecture}

The previous architectures and frameworks discussed have focused primarily on widely studied languages such as English, Indonesian, Chinese, French, and others. Although these languages benefit from extensive resources and established linguistic frameworks, our work diverges by addressing African Bantu languages, which are linguistically distinct and underrepresented in computational research. Bantu languages exhibit unique structural and morphological characteristics, requiring specialized approaches that go beyond the methodologies applied to more commonly studied languages. Hence, the embedding models and the architectural complexities in the components of previous frameworks do not conform to the requirements of the Bantu languages.

Taking LiDA as our base framework, we propose our modified architecture in ~\ref{subsubsec:proposed_arch} creating a robust architecture that caters to the demands of Bantu languages.

\subsubsection{LiDA Architecture}
\begin{figure}[!htbp]
  \includegraphics[width=\columnwidth]{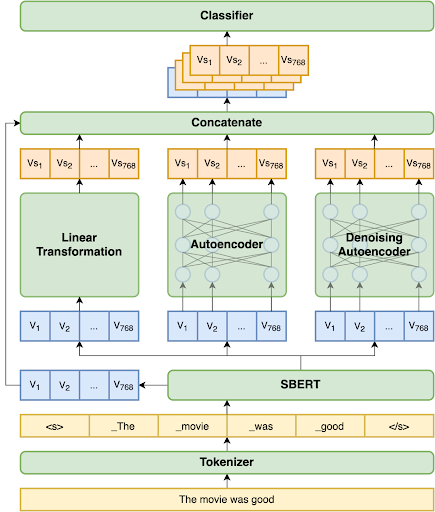}
  \caption{LiDA Framework Reproduced from  \citeauthor{LiDA} \citeyear{LiDA}.}
  \label{fig:methodology}
\end{figure}

The LiDA architecture (\textit{figure~\ref{fig:methodology}}) makes use of the multilingual SBERT model making the architecture \textit{language independent} in essence. The embeddings so generated are passed through three functions - linear transformation, autoencoder model, denoising autoencoder model - before being concatenated with the original embeddings and henceforth classified using LSTM and BERT classifiers.

\subsubsection{Proposed Architecture}
\label{subsubsec:proposed_arch}

\begin{figure}[!htbp]
  \includegraphics[width=\columnwidth]{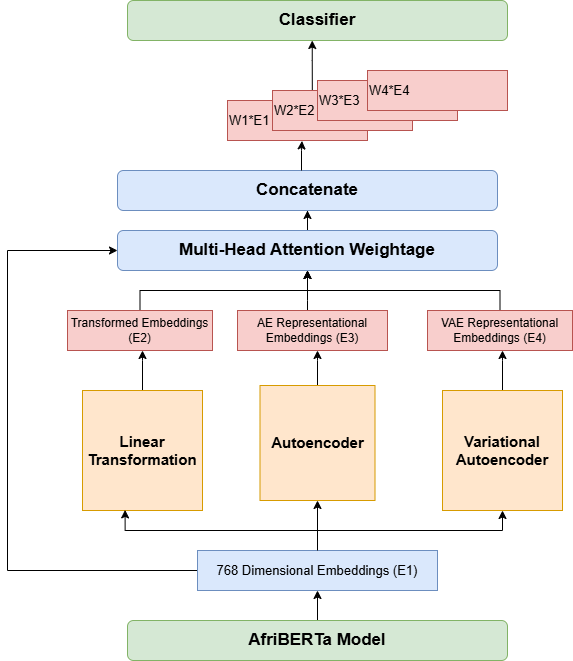}
  \caption{Modified LiDA - MAGE Framework}
  \label{fig:proposed_arch}
\end{figure}

Figure~\ref{fig:proposed_arch} shows our modification to the orignal LiDA architecture by changing the embedding model to AfriBERTa, replacing the denoising autoencoder with a variational autoencoder and addressing the concatenation of complex low-resource languages such as the Bantu language family by the introduction of weighted concatenation using Multi-Head attention.
    The tweets are passed through the AfriBERTa model, the choice of which is discussed in the sub-section~\ref{subsec:afriberta}. The model outputs a 768-dimensional representation of the text which is passed through the aforementioned transformation functions.
    
The Linear Transformation Layer introduces controlled variability into the input embeddings by applying a randomized shift, enhancing the robustness and generalizability of the representations. For each embedding, a random noise vector r is sampled uniformly within a range $[r_{\text{min}}, r_{\text{max}}]$ and added to the original embedding, resulting in a transformed embedding \emph{e' = e+r}. This operation is performed independently for all embeddings in the training and testing datasets. The parameters $r_\text{min}$ and $r_\text{max}$  can be adjusted to control the magnitude of perturbation, ensuring that the embeddings retain their original semantic structure while introducing sufficient variability to aid learning.

Autoencoder is a key part of the augmentation process, designed to refine and diversify input embeddings by learning compressed representations while retaining essential features. This is accomplished through an encoder-decoder architecture that reduces the input embedding dimensions to a latent space and reconstructs them back to the original size. This introduces subtle variations while preserving essential semantic features, enhancing the diversity of augmented data. The model has been slightly enhanced from the original model used in Sujana and Kao (\citeyear{LiDA}) with Leaky ReLU activations, Batch Normalization - to ensure stable training and mitigate vanishing gradients - and Dropout layers to help with regularization, improve generalization and reduce overfitting for the Bantu language family.
The encoder consists of sequential linear layers that progressively reduce the embedding size from the original \emph{768} dimensions to \emph{32} dimensions in the latent space with a learning rate of 0.001 for stabilized learning. The Decoder mirrors the encoder structure, gradually increasing the dimensionality from the latent space back to the original embedding size of \emph{768}, with the final layer applying a sigmoid activation for bounded output. This structure ensures that the embeddings are refined through compression and reconstruction, creating more diverse and robust representations.

Variational Autoencoder proves to be integral to our data augmentation process, providing a more flexible and expressive approach to embedding refinement compared to the original denoising autoencoder. Unlike traditional autoencoders, VAEs model the input data as probabilistic distributions rather than deterministic mappings, allowing for more varied and robust synthetic embeddings. This probabilistic framework facilitates the generation of diverse augmented data points, which is crucial in low-resource language tasks like those involving Bantu languages. The VAE learns a distribution over the latent space, enabling the generation of new samples by sampling from the learned distribution, enhancing the diversity of the training data and helping the model generalize better.
The VAE architecture consists of an encoder and a decoder. The encoder first maps the input embeddings from their original input dimension of \emph{768} to a latent space of dimension \emph{256}. The size of the intermediate layers i.e. hidden dimension is set to 512, balancing capacity and complexity. Batch Normalization, ReLU activations, and Dropout (set to 0.2) are applied to improve stability, avoid overfitting, and ensure robust learning. The encoder then produces two outputs: the mean $\mu$ and log-variance log($\sigma$\textsuperscript{2}) of the latent distribution. The reparameterization trick is applied to sample from this distribution, allowing gradients to propagate through the sampling process and enabling effective training.
Just as in the autoencoder, decoder mirrors the encoder, first mapping the latent representation of latent dimension back to the hidden space, and then expanding to the original input dimension. The final output of the decoder is a probabilistic reconstruction of the original input.
    
    The point to note here is that we are not proposing that VAE will always outperform the original DAE configuration mentioned in the LiDA architecture, but may prove to be comparable or even superior in some scenarios and both configurations will provide us with the desired end results when used along with attention mechanisms.

Finally, the embeddings we get through these transformations are weighted and concatenated with the original embeddings using a Multi-Head attention mechanism section~\ref{sec:mage}.

\subsection{AfriBERTa as the embedding model}
\label{subsec:afriberta}
For this study, we selected the AfriBERTa model by Ogueji et al. (\citeyear{ogueji-etal-2021-small}) model to generate embeddings. This decision was informed by a detailed comparative analysis of multiple models, including BantuBERTa Parvess (\citeyear{parvess2023bantuberta}), afro-xlmr-large by Alabi et al. (\citeyear{alabi-etal-2022-adapting}), and mBERT, across critical classification metrics such as accuracy, precision, recall, and F1-score (see table~\ref{table:mbert_xlmr} and table~\ref{table:afri_bantu}) on both individual fine-tuning for each of the three languages as well as on the combined dataset as used in our final experiments as well. AfriBERTa's results align with the study Thangaraj et al. (\citeyear{thangaraj2024crosslingualtransfermultilingualmodels}) highlighting Afriberta’s effectiveness, particularly in low-resource settings, by achieving the highest performance metrics after fine-tuning in contrast to other comparable models including BantuBERTa. Language-specific fine-tuning results are present in the supplementary material for reference.
\begin{table}[h]
    \centering
    
    \begin{tabular}{lcc}
        \toprule
        \textbf{Metric} & \textbf{mBERT} & \textbf{Afro XLM-R} \\
        \midrule
        Accuracy  & 0.483  & 0.520  \\
        F1 Score  & 0.425  & 0.522  \\
        Precision & 0.464  & 0.528  \\
        Recall    & 0.455  & 0.549  \\
        \bottomrule
    \end{tabular}
    \caption{Performance metrics for mBERT and Afro XLM-R}
    \label{table:mbert_xlmr}

\end{table}

\begin{table}[h!]
    \centering
    
    \begin{tabular}{lcc}
        \toprule
        \textbf{Metric} & \textbf{AfriBERTa} & \textbf{BantuBERTa} \\
        \midrule
        Accuracy  & 0.569  & 0.526 \\
        F1 Score  & 0.580  & 0.513 \\
        Precision & 0.578  & 0.520 \\
        Recall    & 0.560  & 0.511 \\
        \bottomrule
    \end{tabular}
    \caption{Performance metrics for AfriBERTa and BantuBERTa.}
    \label{table:afri_bantu}

\end{table}

\vspace{3cm}

\subsection{MAGE}
\label{sec:mage}

The Multi-Head Attention component is designed to enhance the model's ability to focus on important embeddings by assigning distinct weights to different embeddings using multiple attention heads. This method allows for the dynamic selection of which embeddings have more influence on the final classification decision, effectively highlighting the critical features. The num\_heads was set to \emph{4} ensuring that multiple perspectives of the embeddings are captured simultaneously. The use of attention mechanism stemmed from the observation that on manually weighting embeddings improved classification results were observed, and attention provides a learnable way to optimize this process. In this approach, the embeddings are attended to using Multi-Head attention, where each head independently processes the embeddings and captures different aspects of the feature space. The context vectors, which are trainable, guides the attention mechanism to focus on the most relevant embeddings. The outputs from each attention head are then concatenated, and an aggregation is performed by summing the resulting vectors, capturing the most important features across different heads.

\subsection{Classification}
We employed two architectures for classification, namely, LSTM and Logistic Regression and the results were evaluated using Accuracy, Precision, Recall and F1 Score. The LSTM classifier was used with an input dimension of \emph{768}, hidden dimension of \emph{128}, and a single layer. The model was trained with a learning rate of \emph{0.001} using CrossEntropyLoss. An Early Stopping mechanism with a patience of \emph{3} epochs was employed, along with the StepLR scheduler to adjust the learning rate.

The second classifier used was Logistic Regression, which serves as a lightweight yet effective baseline for classification. The model was trained with a maximum of 1000 iterations using the LBFGS solver for optimization. Since Logistic Regression is a simple linear model, it provides a useful comparison against the LSTM’s sequential feature extraction capabilities. By analyzing both models, we aim to assess the impact of complex sequential modeling versus traditional linear classification on our dataset.

\section{Results}
In this section, we present the classification performance of various embedding configurations and attention mechanisms, evaluated using standard metrics such as accuracy, precision, recall, and F1 score. In addition, a more detailed ablation study is included in the supplementary material, where we also explore grouphead attention as an alternative attention mechanism.

\subsection{Effect of DAE and VAE}

We first compare the performance of the original embeddings using the proposed VAE vs DAE configurations using LSTM and Logistic Regression without weighted attention concatenation mechanism.

\begin{table}[h!]
    \centering
    \begin{tabular}{lccc}
        \toprule
        \textbf{Metric} & \textbf{Original} & \textbf{With DAE} & \textbf{With VAE} \\
        \midrule
        Accuracy  & 0.5680 & 0.5739 & 0.5769 \\
        Precision & 0.5589 & 0.5659 & 0.5681 \\
        Recall    & 0.5554 & 0.5697 & 0.5562 \\
        F1 Score  & 0.5566 & 0.5672 & 0.5589 \\
        \bottomrule
    \end{tabular}
    \caption{Comparison between DAE and VAE configurations (LSTM).}
    \label{tab:results}
\end{table}

Observing table~\ref{tab:results} we see an increase in the metrics over the original results for the DAE and VAE configuration using LSTM classifier. 

\begin{table}[h!]
    \centering
    
    \begin{tabular}{lccc}
        \toprule
        \textbf{Metric} & \textbf{Original} & \textbf{With DAE} & \textbf{With VAE} \\
        \midrule
        Accuracy  & 0.5710 & 0.5729 & 0.5735 \\
        Precision & 0.5636 & 0.5672 & 0.5659 \\
        Recall    & 0.5596 & 0.5624 & 0.5621 \\
        F1 Score  & 0.5592 & 0.5619 & 0.5618 \\
        \bottomrule
    \end{tabular}
    \caption{Comparison between DAE and VAE configurations (Logistic Regression).}
    \label{tab:results_2}

\end{table}

\begin{figure*}[t]
  \centering
  \includegraphics[scale=0.5]{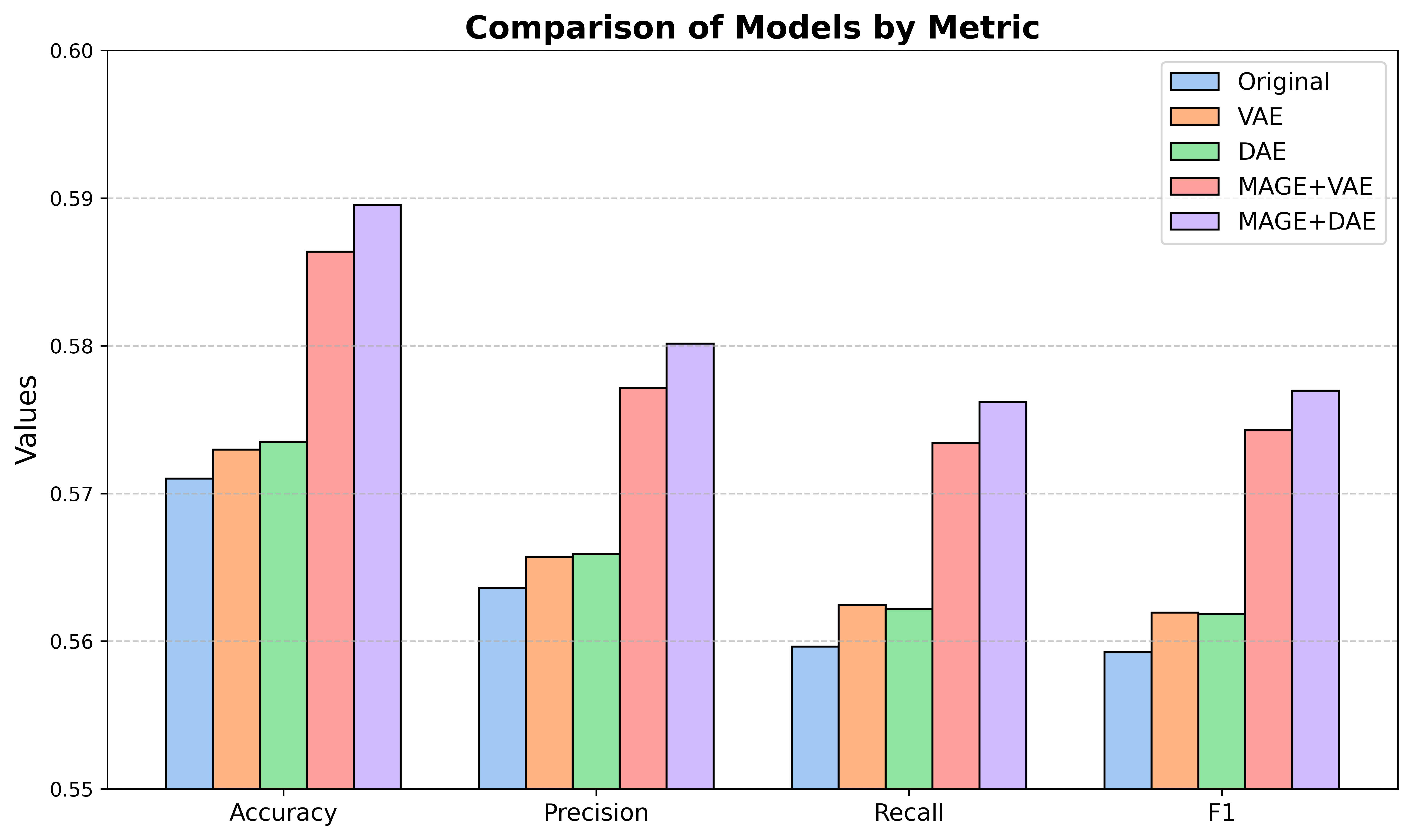}
  \caption{Comparative Results}
  \label{graph:lr}
\end{figure*}

Similarly for logistic regression, in table~\ref{tab:results_2}, we observe for VAE configuration - 0.34\%, 0.37\%, 0.50\%, 0.48\% improvement in accuracy, precision, recall and F1 score respectively over the original classification results and similarly for DAE configuration at 0.43\%, 0.40\%, 0.45\% and 0.46\% improvement for the aforementioned metrics. Both LSTM and Logistic Regression provide us with similar results. One observation to point out here is the comparable results of DAE and VAE configurations, which presents us with an interchangeable choice between both of the configurations.

\subsection{Integrating Multi-Head  Attention}

\begin{figure}[t]
  \centering
  \includegraphics[scale=0.45]{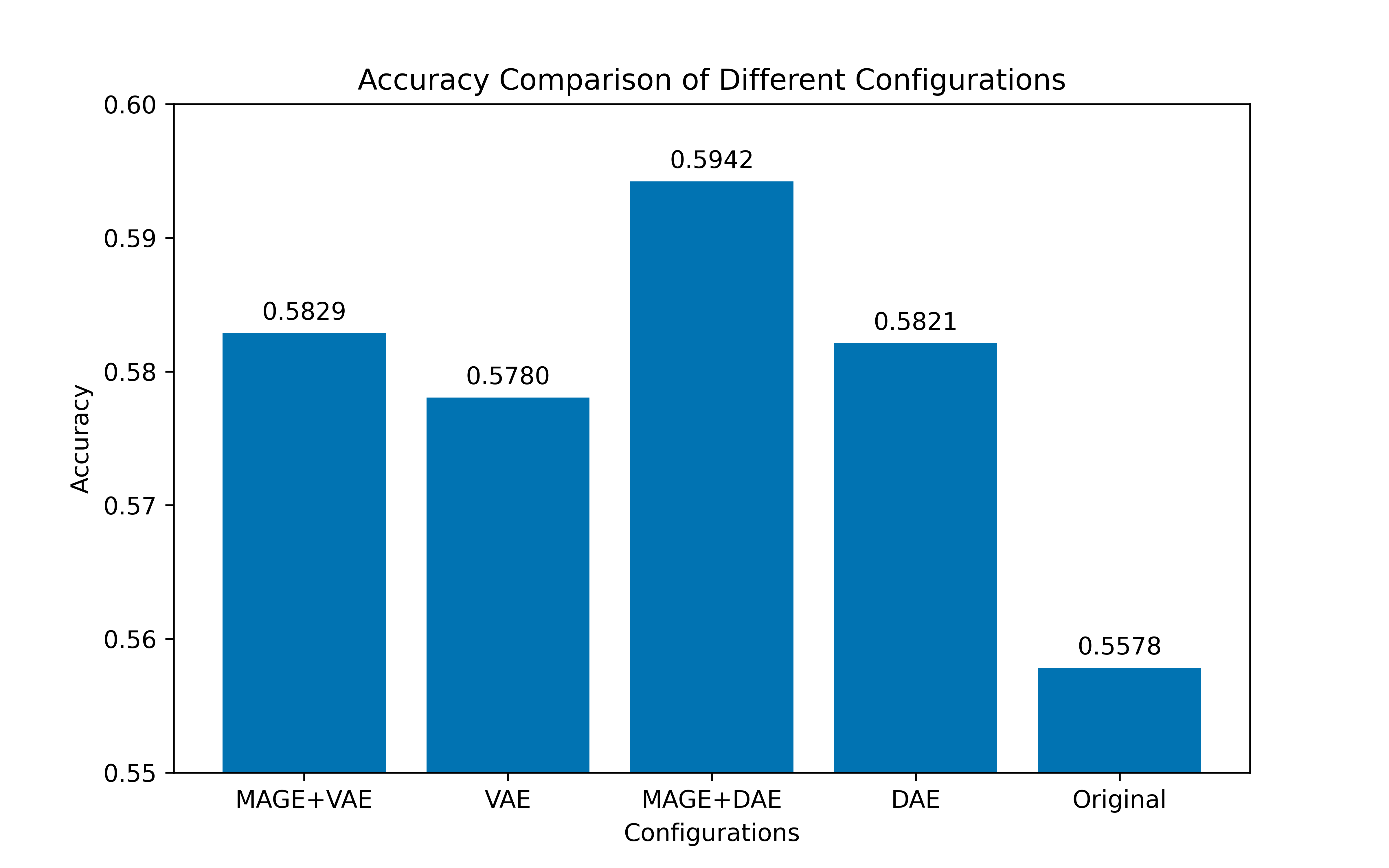}
  \caption{Comparative Results}
  \label{graph:lstm}
\end{figure}

For greater generalization and to test the reliability of attention mechanisms we benchmarked the LSTM classifier by shuffling the dataset 4 times, in each shuffle running the benchmarks for 5 stand-alone iterations. Figure~\ref{graph:lstm} visualizes the average accuracy achieved over the shuffles comparing with and without Multi-Head  attention with both DAE and VAE configurations. 
The drastic improvement in classification accuracy compared to the baseline embeddings are clearly visualized here with the highest increase of 3.64\% for MAGE+DAE over the baseline,  closely followed by MAGE+VAE at 2.51\% improvement. 
The attention mechanism further boosts the VAE and DAE configurations with an increase of 0.4\% and 1.21\% respectively.

Similarly, we followed a similar shuffling trend for benchmarking the logistic regression classifier achieving the averaged out results as noted in figure~\ref{graph:lr}. We can observe how our attention configurations outperform the baseline configurations across all metrics.



\section{Conclusion}

We thus present an innovative approach to embedding refinement and classification for the Bantu language family by integrating embedding level transformations and advanced attention mechanisms. Through systematic experimentation, we demonstrated that denoising and variational autoencoders enhance the quality of embeddings by refining their structure while preserving semantic integrity.
A key contribution of this study was the introduction of a Multi-Head Attention mechanism. The attention mechanism dynamically assigns different weights to various embeddings, enabling the model to focus on the most relevant features. This approach allowed us to capture crucial aspects of the embedding space more effectively. The results from our experiments clearly indicate that Multi-Head Attention significantly boosted classification performance across various metrics. This validates the hypothesis that refining embeddings and emphasizing important features through attention enhances classification performance. The framework proposed in this study provides a scalable and robust solution, particularly for low-resource languages where linguistic diversity and data scarcity pose unique challenges.


\section{Limitations}
While our proposed approach demonstrates notable improvements in embedding refinement and classification, it has several limitations. First, our experiments were conducted on a limited set of three Bantu languages which restricts the generalizability of our findings to other Bantu and low-resource languages. Second, the dataset exhibits an imbalance, with Kinyarwanda comprising the majority of data points. This skewness may introduce biases in model learning and affect the performance across languages. Third, the dataset size may be insufficient for training complex components like the Denoising Autoencoder, Variational Autoencoder, and the standard Autoencoder as they require a large and diverse dataset to learn meaningful latent representations effectively. The limited training data could lead to suboptimal embeddings, affecting downstream classification performance. Fourth, while embedding-level transformations using denoising and variational autoencoders refine embedding structures, their impact on preserving linguistic nuances requires further investigation. Finally, the computational complexity of our approach, particularly the attention mechanism, may pose challenges for real-time applications in resource-constrained environments. Addressing these limitations in future research will be essential for broader adoption and scalability.

\section*{Acknowledgments}
We would like to express our gratitude to our mentor Dr. Vukosi Marivate for his guidance and support, as well as Data Science for Social Impact (DSFSI) Lab of University of Pretoria for providing valuable research insights and resources. The code for this work catchat
n be accessed at \href{https://github.com/dsfsi/gdev-2024/tree/master}{Repository}.

\bibliography{main}

\end{document}